\documentclass[sn-mathphys]{sn-jnl}

\usepackage{import}
\usepackage{amsmath,graphicx}
\usepackage{times}
\usepackage[dvipsnames]{xcolor}

\usepackage{tabularx,booktabs}
\usepackage{arydshln}
\usepackage{tabu}
\usepackage{adjustbox}
\usepackage{enumerate,multicol}
\usepackage{multirow}
\usepackage{multicol}

\usepackage{gensymb}
\usepackage{float}
\usepackage{balance}
\usepackage{xspace}
\usepackage{amssymb}
\usepackage{bbding,enumitem}
\usepackage{booktabs}
\usepackage[pagebackref]{hyperref}
\usepackage{color, colortbl}
\hypersetup{
    colorlinks,
    linkcolor={red!50!black},
    citecolor={blue!50!black},
    urlcolor={black}
}

\usepackage{times}

\usepackage{soul}
\usepackage{url}

\usepackage{amsmath}
\usepackage{nicefrac, xfrac}
\usepackage{booktabs}

\definecolor{RYB2}{RGB}{245,245,245}
\definecolor{RYB1}{RGB}{218,232,252}
\definecolor{RYB4}{RGB}{108,142,191}
\definecolor{shadecolor1}{rgb}{0.95,0.95,0.95}

\newcommand{\avsum}{\mathop{\mathpalette\avsuminner\relax}\displaylimits}
\makeatletter
\newcommand\avsuminner[2]{%
  {\sbox0{$\m@th#1\sum$}%
   \vphantom{\usebox0}%
   \ooalign{%
     \hidewidth
     \smash{\vrule height\dimexpr\ht0+1pt\relax depth\dimexpr\dp0+1pt\relax}%
     \hidewidth\cr
     $\m@th#1\sum$\cr
   }%
  }%
}

\usepackage{verbatim}

\jyear{2023}

\theoremstyle{thmstyleone}

\theoremstyle{thmstyletwo}

\theoremstyle{thmstylethree}

\newcommand{\ie}{\textit{i.e.}, }

\title[DeepPyramid+]{DeepPyramid+: Medical Image Segmentation using Pyramid View Fusion and Deformable Pyramid Reception\footnote{This work was funded by Haag-Streit Foundation, Switzerland.}}

\author*[1]{\fnm{Negin} \sur{Ghamsarian }}\email{negin.ghamsarian@unibe.ch}

\author[3]{\fnm {Sebastian} \sur{Wolf}\email{sebastian.wolf@insel.ch}}

\author[3]{\fnm{Martin} \sur{Zinkernagel}\email{martin.zinkernagel@insel.ch}}

\author[2]{\fnm{Klaus} \sur{Schoeffmann}}\email{ks@itec.aau.at}

\author[1]{\fnm{Raphael} \sur{Sznitman}}\email{raphael.sznitman@unibe.ch}

\affil[1]{\orgdiv{ARTORG Center for Biomedical Engineering Research}, \orgaddress{\orgname{University of Bern}, \city{Bern}, \country{Switzerland}}}

\affil[2]{\orgdiv{Department of Information Technology}, \orgaddress{\orgname{University of Klagenfurt}, \city{Klagenfurt}, \country{Austria}}}

\affil[3]{\orgdiv{Department of Ophthalmology}, \orgaddress{\orgname{Inselspital}, \city{Bern}, \country{Switzerland}}}

\begin{document}

\abstract{

\textbf{Purpose} Semantic Segmentation plays a pivotal role in many applications related to medical image and video analysis. However, designing a neural network architecture for medical image and surgical video segmentation is challenging due to the diverse features of relevant classes, including heterogeneity, deformability, transparency, blunt boundaries, and various distortions. We propose a network architecture, DeepPyramid+, which addresses diverse challenges encountered in medical image and surgical video segmentation.\\
\textbf{Methods}   The proposed DeepPyramid+ incorporates two major modules, namely ``Pyramid View Fusion" (PVF) and ``Deformable Pyramid Reception," (DPR), to address the outlined challenges. PVF replicates a deduction process within the neural network, aligning with the human visual system, thereby enhancing the representation of relative information at each pixel position. Complementarily, DPR introduces shape- and scale-adaptive feature extraction techniques using dilated deformable convolutions, enhancing accuracy and robustness in handling heterogeneous classes and deformable shapes.\\
\textbf{Results} Extensive experiments conducted on diverse datasets, including endometriosis videos, MRI images, OCT scans, and cataract and laparoscopy videos, demonstrate the effectiveness of DeepPyramid+ in handling various challenges such as shape and scale variation, reflection, and blur degradation. DeepPyramid+ demonstrates significant improvements in segmentation performance, achieving up to a 3.65\% increase in Dice coefficient for intra-domain segmentation and up to a 17\% increase in Dice coefficient for cross-domain segmentation.\\
\textbf{Conclusions} DeepPyramid+ consistently outperforms state-of-the-art networks across diverse modalities considering different backbone networks, showcasing its versatility. Accordingly, DeepPyramid+ emerges as a robust and effective solution, successfully overcoming the intricate challenges associated with relevant content segmentation in medical images and surgical videos. Its consistent performance and adaptability indicate its potential to enhance precision in computerized medical image and surgical video analysis applications.
}

\keywords{Semantic Segmentation, Medical Images, Surgical Videos, Neural Networks, Deformable Convolutions, Dilated Convolutions, Pyramid View Fusion, Deformable Pyramid Reception}
\maketitle

\section{Introduction}
\label{sec: introduction}

Semantic segmentation has emerged as a critical tool in computerized medical image and surgical video analysis, empowering numerous applications in various domains. In surgical videos, semantic segmentation is a prerequisite in several applications ranging from  phase and action recognition, irregularity detection, surgical training, objective skill assessment, relevance-based compression, surgical planning, operation room organization, and so forth~\cite{RDC,RBE,RelComp,Adapt-Net}. In the case of volumetric medical images, semantic segmentation can considerably aid in the diagnosis, treatment planning, and monitoring~\cite{AIDS}. 
Automatic segmentation of medical images and videos can also reduce subjective errors caused by time constraints and workloads while enhancing treatment and surgical efficiency.

Designing a neural network architecture for medical image and surgical video segmentation presents a challenge due to the diverse features exhibited by different relevant labels. Specifically, many classes of objects relevant to the medical image and surgical video analysis are heterogenous, featuring deformable or amorphous instances, as well as color,  texture, and scale variation. Besides in surgical videos, the problem of motion blur degradation becomes more critical due to the camera's proximity to the surgical scene. Unlike general images, medical images and surgical videos may contain transparent relevant content (such as intraocular lens) or exhibit blunt boundaries, further complicating the task of semantic segmentation. Accordingly, an effective network for medical image and surgical video segmentation should be able to simultaneously deal with (I) heterogeneity and deformability in relevant objects, and (II) transparency, blunt edges, and distortions such as motion and defocus blur. 

This paper introduces a U-Net-based CNN for semantic segmentation, which effectively addresses the challenges associated with segmenting relevant content in medical images and surgical videos by adaptively capturing semantic information\footnote{This paper is an extended version of DeepPyramid~\cite{DeepPyramid}, featuring minor enhancements in the DPR module.}. The proposed network, called DeepPyramid+, comprises two key modules: (i) Pyramid View Fusion (PVF) module, which offers a narrow-to-wide-angle global view of the feature map centering at each pixel position, and (ii) Deformable Pyramid Reception (DPR) module, responsible for performing shape-adaptive feature extraction on the input convolutional feature map. 
We provide comprehensive experiments to compare the performance of DeepPyramid+ with state-of-the-art baselines for five intra-domain and two cross-domain datasets. Experimental results reveal the superiority of DeepPyramid+ compared to the baselines. Ablation studies confirm the effectiveness of each proposed module in boosting semantic segmentation performance. To support reproducibility and further investigations, we will release the PyTorch implementation of DeepPyramid+ and all dataset splits with the acceptance of this paper.

\section{Related Work}
\label{sec: relatedwork}

U-Net~\cite{U-Net} was initially proposed for medical image segmentation and achieved succeeding performance being attributed to its skip connections. Many U-Net-based architectures have been proposed over the past years to improve the segmentation accuracy and address different flaws and restrictions in the previous architectures~\cite{FED-Net, RAUNet, CE-Net, PAANet, BARNet, CPFNet, UNet++}. 

\vspace{0.5\baselineskip}
\noindent{\textit{\textbf{Attention Modules. }}}
Attention mechanisms can be broadly described as the techniques to guide the network's computational resources (\ie the convolutional operations) toward the most determinative features in the input feature map~\cite{SCSE, RAUNet, ReCal-Net}. Such mechanisms have been especially proven to be gainful in the case of semantic segmentation.  
The scSE blocks~\cite{SCSE} aim to recalibrate the feature maps based on pixel-wise and channel-wise global features.
BARNet~\cite{BARNet} adopts a bilinear-attention module to extract the cross dependencies between the different channels of a convolutional feature map. 
PAANET~\cite{PAANet} uses a double attention module to model semantic dependencies between channels and spatial positions in the convolutional feature map.

\vspace{0.5\baselineskip}
\noindent{\textit{\textbf{Fusion Modules. }}}
Fusion modules can be characterized as modules designed to improve semantic representation via combining several feature maps. The input feature maps could range from varying-level semantic features to the features coming from parallel operations.
PSPNet~\cite{PSPNet} adopts a pyramid pooling module (PPM) containing parallel sub-region average pooling layers followed by upsampling to fuse the multi-scale sub-region representations. 
Atrous spatial pyramid pooling (ASPP)~\cite{DeepLab, DeepLabv3+} was proposed to deal with objects' scale variance by aggregating multi-scale features extracted using parallel varying-rate dilated convolutions.
CPFNet~\cite{CPFNet} uses another fusion approach for scale-aware feature extraction.

\section{Methodology}
\label{sec: Methodology}

\begin{figure}[!tb]
    \centering
    \includegraphics[width=1\textwidth]{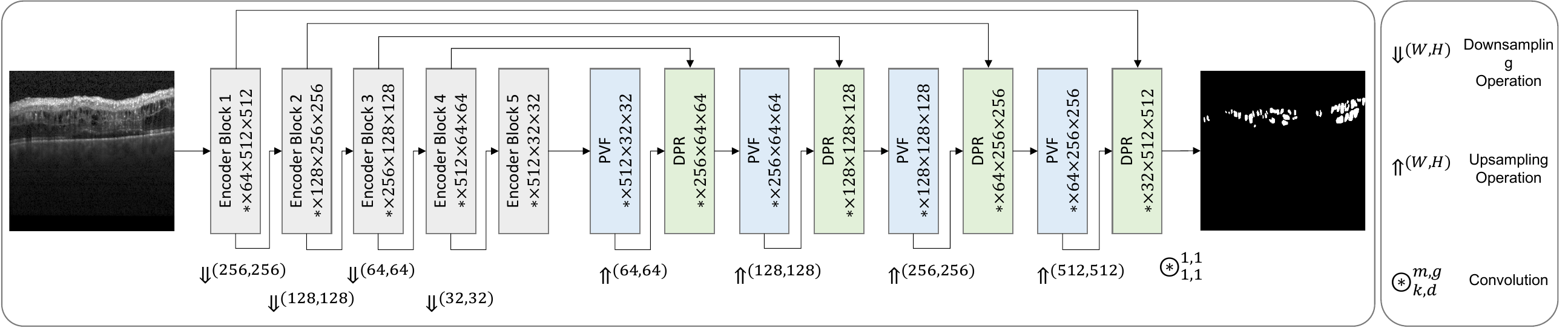}
    \caption{Overall architecture of DeepPyramid+ consisting of encoder blocks of the VGG16 network, and the proposed PVF and DPR modules. The numbers in each block correspond to the output feature map's dimensions.}
    \label{fig:BD}
\end{figure}

We present a segmentation network that focuses on (I) modeling heterogeneous classes featuring deformations, shape, scale, color, and context variation, (II) dealing with content distortion due to motion blur and reflection, and (III) handling objects' transparency and blunt boundaries (Fig.~\ref{fig:BD}). 
At its core, our network adopts the U-Net architecture, with the encoder part being set to VGG16.
We develop two decoder modules specifically tailored to tackle the mentioned challenges:
(1) \textit{Pyramid View Fusion (PVF)}, which aims to replicate a deduction process within the neural network analogous to the functioning of the human visual system by enhancing the representation of relative information at each individual pixel position.
(2) \textit{Deformable Pyramid Reception (DPR)},
which addresses the limitations of regular convolutional layers by introducing deformable dilated convolutions and shape- and scale-adaptive feature extraction techniques. This module allows for handling the complexities of heterogeneous classes and deformable shapes, resulting in improved accuracy and robustness in the segmentation performance.

\noindent{We specify the functionality of each module in the following subsections. Additional discussions regarding the effectiveness of each module and an analysis of the complexity for each module are available in the supplementary material.}

\vspace{0.5\baselineskip}
\noindent\textbf{\textit{Notations. }}Throughout this paper, we represent convolutional layers with a kernel size of $(k\times k)$, dilation of $d$, $m$ output channels, and $g$ groups as $\circledast_{k,d}^{m,g}$. For deformable convolutions, we use the symbol ${\tilde{\circledast}}_{k,d}^{m,g}$. {Additionally, we illustrate} the average-pooling layer with a kernel size of $(k\times k)$ and a stride of $s$ pixels as $\avsum_{k}^{s}$, and global average pooling as $\avsum^{G}$. The symbol $\concat_{D}$ denotes feature map concatenation over dimension $D$. Furthermore, we employ $\Uparrow^{(W_{out}, H_{out})}$ and $\Downarrow^{(W_{out}, H_{out})}$ for upsampling and downsampling operations with a scale factor of $(W_{out}, H_{out})$, respectively. We use $\sigma(\cdot)$ to represent the Softmax operation, $\lVert \cdot \rVert_{n}$ for layer normalization over the last $n$ dimensions, $\mathcal{R}(\cdot)$ for the ReLU nonlinearity function, and $\tau(\cdot)$ for the hard tangent hyperbolic function.

\subsection{Pyramid View Fusion (PVF)} 
\begin{figure}[!tb]
    \centering
    \includegraphics[width=1\textwidth]{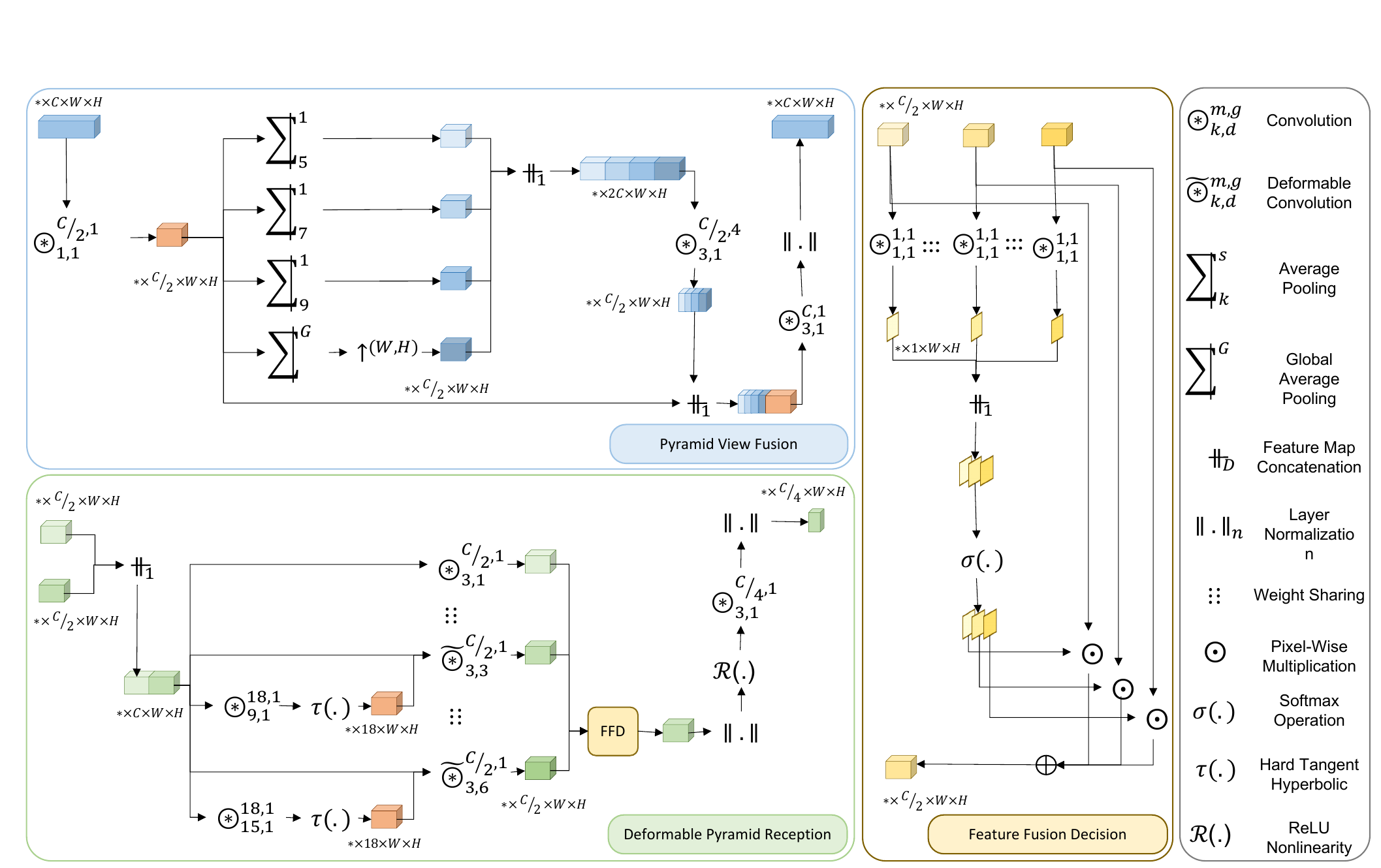}
    \caption{The detailed architecture of the PVF and DPR modules.}
    \label{fig:modules}
\end{figure}

To optimize computational complexity, the initial step involves creating a bottleneck by employing a convolutional layer with a kernel size of one, as illustrated in Fig.~\ref{fig:modules}. Following this dimensionality reduction stage, the resulting convolutional feature map is fed into four parallel branches. The first branch features a global average pooling layer, which is subsequently followed by upsampling.  The other three branches employ average pooling layers with progressively increasing filter sizes while maintaining a stride of one pixel. The use of a one-pixel stride is specifically important to achieve a pixel-wise centralized pyramid view, as opposed to the region-wise pyramid attention approach employed in PSPNet~\cite{PSPNet}. The output feature maps from all branches are then concatenated and fed into a convolutional layer with four groups, for extracting inter-channel dependencies during dimensionality reduction. Subsequently, a regular convolutional layer is applied to extract joint intra-channel and inter-channel dependencies. The resulting feature map is then passed through a layer-normalization function, which helps normalize the activations for improved stability and performance.

\subsection{Deformable Pyramid Reception (DPR)}
The architecture of the Deformable Pyramid Reception (DPR) module, as depicted in Fig.~\ref{fig:modules}, can be described as follows. Initially, the upsampled coarse-grained semantic feature map from the preceding layer is concatenated with its symmetric fine-grained feature map from the encoder.
Subsequently, these concatenated features are passed through three parallel branches. The first branch employs a regular convolution operation  while the other two branches utilize deformable convolutions with different dilation rates of three and six. 
The structured convolution covers the immediate neighboring pixels up to one pixel to the central pixel. The deformable convolutions with the dilation rate of three and six cover an area from two to four, and five to seven pixels far away from each central pixel, respectively. Accordingly, the DPR module forms a learnable sparse receptive field of size $15\times 15$ pixels by incorporating these layers. These layers share the weights to avoid imposing a huge number of trainable parameters. 

To compute the feature-map-adaptive offset field for each deformable convolution, a regular convolution operation is employed. Considering the target area of the two deformable convolutions, the offset field should be computed based on the internal content within four and seven pixels away from each central pixel ($k=9$, $k=15$). The computed offset values are then passed through a tangent hyperbolic function, which clips them within the range of $[-1, 1]$,  to ensure that each deformable convolution adaptively covers an area within the range of $[k-1, k+1]$.
The offset field provides two values per element in the deformable convolutional kernel (horizontal and vertical offsets). Accordingly, the number of offset field's output channels for a deformable convolution with a kernel of size $3\times 3$ is equal to 18. This enables the deformable convolution to spatially adjust its receptive field based on the learned offset values, improving its ability to capture contextually relevant information.

The output feature maps of the parallel structured and deformable convolutions are then passed through a feature fusion decision (FFD) module~\cite{Adapt-Net}. This module determines the significance of each input feature map based on the spatial descriptors using pixel-wise convolutions. These descriptors are concatenated and subjected to a Softmax operation, resulting in normalized descriptors. 
The normalized descriptors determine the pixel-wise contribution or weight of each input convolutional feature map in the final fused feature map. 
The output feature map of the FFD module is obtained as a weighted sum of the input feature maps, where the normalized descriptors serve as pixel-wise weights. The resulting feature map from the FFD module goes through a series of additional operations for deeper feature extraction and normalization.

\section{Experimental settings}
\label{sec: experimental settings}

\begin{table}[bt!]
\centering
\caption{Specifications of the single-domain and cross-domain datasets.}
\label{table:dataset}
\resizebox{0.8\textwidth}{!}{%
\begin{tabular}{l*{2}{>{\centering\arraybackslash}m{1.8cm}}
*{1}{>{\centering\arraybackslash}m{3cm}}
*{3}{>{\centering\arraybackslash}m{1.8cm}}}
\toprule
 & Application & Modality & Objects & Folds & Train$\vert$Test Size (per Fold) & Reference\\ \cmidrule(lr){2-7}

\multirow{5}{*}{Single Domain}&Cataract & Video & Instruments & 4 & 207$\vert$138 & Cataract-1K\footnote{This dataset is a small subset of Cataract-1K dataset that will be released upon the acceptance of this paper.}\\
&Laparoscopy & Video & Instruments & 4 & 109$\vert$1179 & Endovis~\cite{Endovis2015}\\
&Endometriosis & Video & Endometrial Implant & 4 & 119$\vert$39 & ENID~\cite{ENID}\\
&Prostate MR & MRI & Prostate & 4 & 275$\vert$110 & MS-Net~\cite{MS-Net} \\
&Retina & OCT & IRF Fluid & 4 & 105$\vert$299 & RETOUCH~\cite{RETOUCH} (Spectralis)\\

\bottomrule
\end{tabular}
}

\resizebox{0.8\textwidth}{!}{%
\begin{tabular}{l*{7}{>{\centering\arraybackslash}m{2cm}}}
\toprule
& Application & Modality & Objects & Source$\vert$Target Set & Folds & Train$\vert$Test Size (per Fold) & Reference\\ \cmidrule(lr){2-8}
\multirow{2}{*}{Cross Domain} &Cataract & Video & Instruments & Cataract-1K$\vert$CaDIS & 4 & 207$\vert$458 & {Cataract-1K}$\vert$CaDIS~\cite{CaDIS}\\
&Prostate MR & MRI & Prostate & BMC$\vert$BIDMC & 4 & 275$\vert$64 & MS-Net~\cite{MS-Net} \\

\bottomrule
\end{tabular}
}
\end{table}

\begin{figure}[!tb]
\centering\includegraphics[width=0.8\textwidth]{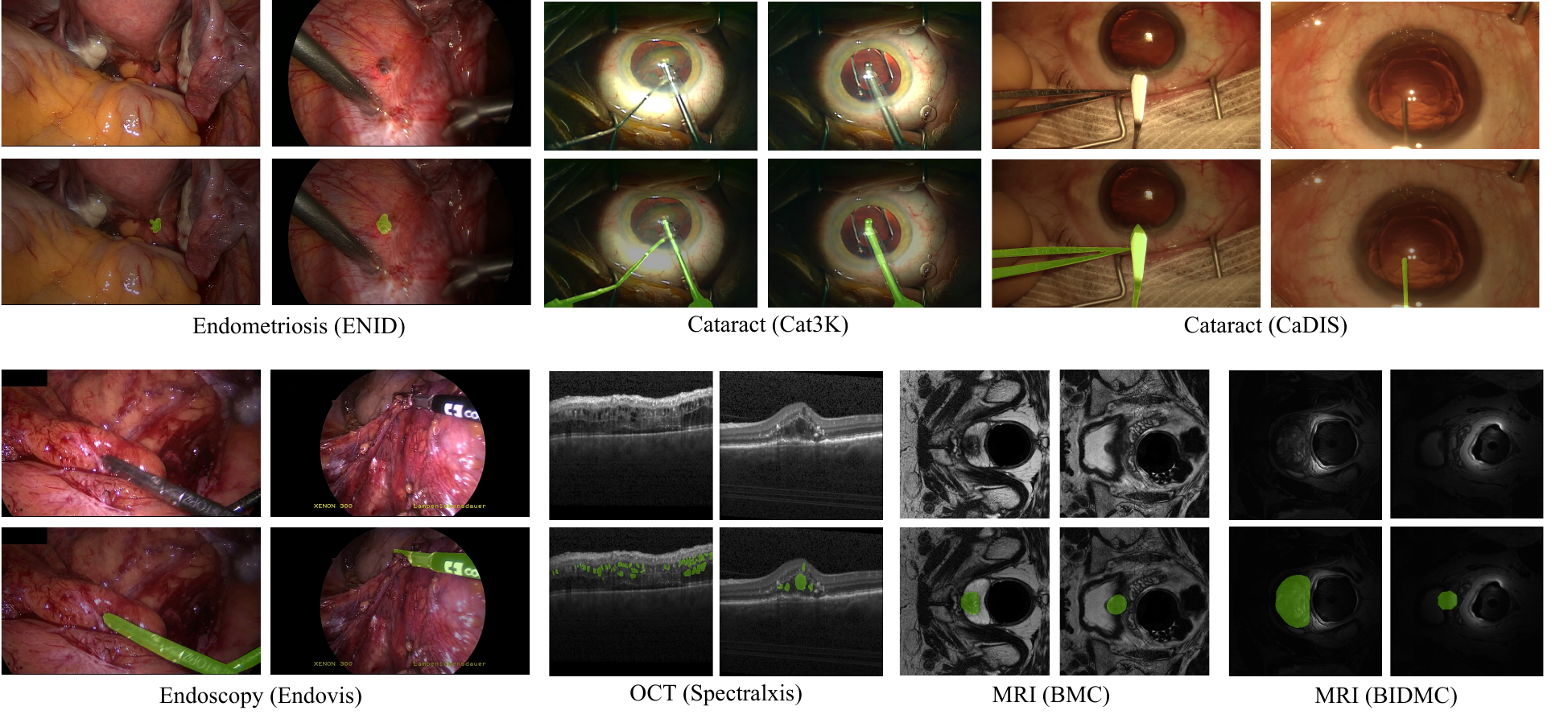}
    \caption{Exemplary images from the different datasets along with their corresponding overlayed masks.}
    \label{fig:datasets}
\end{figure}

\vspace{0.5\baselineskip}
\noindent\textbf{\textit{Datasets. }}We evaluate the performance of our proposed network on five intra-domain datasets from three different modalities (video, MRI, and OCT) and two cross-domain datasets from two different modalities. Table~\ref{table:dataset} details the specifications of adopted datasets and Fig.~\ref{fig:datasets} presents exemplary images together with the ground-truth segmentations from each dataset. 
These datasets cover a wide range of object classes with distinct characteristics. For example, endometriosis videos contain amorphous endometrial implants with color and texture variations. OCT scans involve amorphous intraretinal fluid, while prostate MRI images include deformations and variations in scale, contrast, and brightness.
In addition, instrument segmentation in cataract and laparoscopy surgeries presents various challenges, such as scale variation, reflection, motion blur, and defocus blur degradation.
The diversities in datasets ensure realistic conditions for evaluating the proposed network's effectiveness in addressing challenges in medical image and surgical video segmentation \footnote{This paper aims to design a dedicated network tailored to address medical image and video segmentation challenges, emphasizing various modalities but not within a multi-modal training framework. We substantiate the efficacy of our model through distinctive validations across diverse medical image and video datasets.}. For result reproducibility, we provide all train/test sets as CSV files in the paper's GitHub repository.

\vspace{0.5\baselineskip}
\noindent\textbf{\textit{Alternative Methods. }} We compare the effectiveness of our proposed network architecture with eleven state-of-the-art neural networks using different backbones.  Table~\ref{tab:alternatives}  lists the specifications of the baselines and the proposed network. Note that UNet+ is an improved version of UNet, where we use VGG16 as the backbone network and double convolutional blocks (two consecutive convolutions followed by batch normalization and ReLU layers) as decoder modules. To have fair comparisons with alternative methods, we report the performance of DeepPyramid+ with three different backbones (VGG16, ResNet34, and ResNet50). 

\vspace{0.5\baselineskip}
\noindent\textbf{\textit{Training Settings. }}
All backbones are initialized with the ImageNet pre-trained parameters. We use a batch size of four for all datasets, set the initial learning rate to 0.001, and decrease it during training using polynomial decay $lr = lr_{init}\times (1-\frac{iter}{total-iter})^{0.9}$. The input size of the networks is $512\times 512$ for all datasets. We apply cropping and random rotation (up to 30 degrees), color jittering (brightness = 0.7, contrast = 0.7, saturation = 0.7), Gaussian blurring, and random sharpening as augmentations during training, and use the \textit{cross entropy log dice} loss during training \cite{DeepPyramid}. All experiments are conducted using NVIDIA RTX:3090 GPUs.

\begin{table}[tb!]
\renewcommand{\arraystretch}{1}
\caption{Specifications of the proposed and alternative approaches. In ``Upsampling" column, ``Trans Conv" stands for \textit{Transposed Convolution}.}

\label{tab:alternatives}
\centering
\resizebox{0.85\textwidth}{!}{%
\begin{tabular}{lccccc}
\specialrule{.12em}{.05em}{.05em}%\tabucline[1pt]{1-4}
Model & Backbone & Params. & Upsampling & Target & Reference\\\specialrule{.12em}{.05em}{.05em}%ine
UNet$++$~&VGG16&24.24 M& Bilinear & Medical Images & \cite{UNet++}\\
CPFNet & VGG16 $\vert$ ResNet34 & 39.17 M $\vert$ 34.66 M& Bilinear & Medical Images & \cite{CPFNet}\\
BARNet&ResNet34&24.90 M& Bilinear & Surgical
Instruments  & \cite{BARNet}\\
PAANet &ResNet34& 22.43M & Trans Conv \& Bilinear & Surgical Instruments  & \cite{PAANet}\\
CE-Net &VGG16 $\vert$ ResNet34& 33.50 M $\vert$ 29.90 M& Trans Conv & Medical Images &  \cite{CE-Net}\\
RAUNet&ResNet34&22.14 M&Trans Conv&  Surgical  Instruments  & \cite{RAUNet}\\
FED-Net&ResNet50&59.52 M& Trans Conv \& PixelShuffle & Liver Lesion & \cite{FED-Net}\\
scSENet & VGG16 $\vert$ ResNet34& 22.90 M $\vert$ 25.25 M& Bilinear & Medical Images & \cite{SCSE}\\
DeepLabV3+& ResNet50& 26.68 M&Bilinear&Scene& \cite{DeepLabV3}\\
UPerNet & ResNet50& 51.26 M&Bilinear&Scene& \cite{UPerNet}\\
U-Net+& VGG16 &22.55 M& Bilinear & Medical Images & \cite{U-Net}\\\midrule
DeepPyramid &VGG16 & 33.57 M& Bilinear & Medical Images & Proposed\\
\specialrule{.12em}{.05em}{0.05em}
\end{tabular}
}
\end{table}

\vspace{0.5\baselineskip}
\noindent\textbf{\textit{Ablation Study Settings. }}
To evaluate the effectiveness of different modules, we use the improved version of UNet (UNet+), with the same backbone (VGG16) as our baseline. This network does not include any PVF modules. Besides, the DPR module is replaced with a sequence of two convolutional layers, each of which being followed by a batch normalization layer and a ReLU activation.

\section{Experimental results}
\label{sec: experimental results}

\begin{table}[bt!]
\centering
\caption{Quantitative comparisons among the performance of DeepPyramid+ and alternative methods in organ and disease segmentation, with top two results shown in \textcolor{ForestGreen}{green} and \textcolor{Blue}{blue}, respectively.}
\label{tab:quantitative-disease}
\resizebox{0.93\textwidth}{!}{%

\begin{tabular}{lm{1.3cm}*{8}{>{\centering\arraybackslash}m{1.1cm}}}
\specialrule{.12em}{.05em}{.05em}
Modality && \multicolumn{2}{c}{\footnotesize{Endometriosis Surgery}} & \multicolumn{2}{c}{\footnotesize{MRI}} & \multicolumn{2}{c}{\footnotesize{OCT}} & \multicolumn{1}{l}{\multirow{2}{*}{\footnotesize{Avg. IoU (\%)}}} \\ \cmidrule(lr){3-4}\cmidrule(lr){5-6}\cmidrule(lr){7-8}
Backbone &  Network & IoU (\%) & Dice (\%) & IoU (\%) & Dice (\%) & IoU (\%) & Dice (\%) \\ \specialrule{.12em}{.05em}{.05em}
\multirow{7}{*}{VGG16} & UNet+ & 51.02 & 64.94 & 72.44 & 82.30 & \textcolor{Blue}{51.89} & \textcolor{Blue}{64.95} & \textcolor{Blue}{58.45}\\
 & scSENet & 48.95 & 62.95 & 72.31 & 82.23 & \textcolor{ForestGreen}{52.12} & \textcolor{ForestGreen}{65.18} & 57.79 \\
 & FEDNet & 50.38 & 64.16 & 71.03 & 81.82 & 48.70 & 62.31 & 56.70\\
 & CE-Net & 44.40 & 58.48 & 70.84 & 81.27 & 47.33 & 61.10 & 54.19\\
 & CPFNet & \textcolor{Blue}{52.82} & \textcolor{Blue}{66.09} & \textcolor{Blue}{73.28} & \textcolor{Blue}{82.80} & 47.90 & 61.17 & 58.00\\
 & UNetPP & 51.02 & 64.83 & 72.49 & 82.32 & 51.58 & 64.60 & 58.36\\
 & DeepPyramid+ & \textcolor{ForestGreen}{53.22} & \textcolor{ForestGreen}{66.37} & \textcolor{ForestGreen}{76.02} & \textcolor{ForestGreen}{85.36} & 50.79 & 63.99 & \textcolor{ForestGreen}{60.01}\\ \midrule
\multirow{5}{*}{ResNet34} 
         & scSENet & \textcolor{Blue}{46.20} & \textcolor{Blue}{60.65} & \textcolor{Blue}{72.99} & \textcolor{Blue}{82.66} & \textcolor{Blue}{49.64} & \textcolor{Blue}{63.31} & \textcolor{Blue}{56.28}\\
 & FEDNet & 28.19 & 37.85 & 70.24 & 80.61 & 45.34 & 59.79 & 47.92\\
 & CE-Net & 11.02 & 18.94 & 13.76 & 17.51 & 15.68 & 25.53 & 13.49\\
 & CPFNet & 23.07 & 35.54 & 59.95 & 73.26 & 18.29 & 28.99 & 33.77 \\
 & DeepPyramid+ & \textcolor{ForestGreen}{54.04} & \textcolor{ForestGreen}{67.63} & \textcolor{ForestGreen}{73.46} & \textcolor{ForestGreen}{83.66} & \textcolor{ForestGreen}{51.22} & \textcolor{ForestGreen}{64.82} & \textcolor{ForestGreen}{59.57}\\ \midrule
\multirow{3}{*}{ResNet50}
 & UPerNet & \textcolor{Blue}{48.93} & \textcolor{Blue}{62.56} & \textcolor{Blue}{72.73} & \textcolor{Blue}{82.68} & \textcolor{Blue}{46.17} & \textcolor{Blue}{60.18} & \textcolor{Blue}{55.94}\\
 & DeepLabV3+ & 43.00 & 56.66 & 70.83 & 80.79 & 44.64 & 58.86 & 52.82\\
 & DeepPyramid+ & \textcolor{ForestGreen}{53.11} & \textcolor{ForestGreen}{67.12} & \textcolor{ForestGreen}{73.93} & \textcolor{ForestGreen}{83.97} & \textcolor{ForestGreen}{50.99} & \textcolor{ForestGreen}{64.63} & \textcolor{ForestGreen}{59.34}\\

\specialrule{.12em}{.05em}{.05em}
\end{tabular}
}
\end{table}

\begin{table}[bt!]
\centering
\caption{Quantitative comparisons among the performance of DeepPyramid+ and alternative methods in instrument segmentation, with top two results shown in \textcolor{ForestGreen}{green} and \textcolor{Blue}{blue}, respectively.}
\label{tab:quantitative-instruments}
\resizebox{0.72\textwidth}{!}{%

\begin{tabular}{lm{1.3cm}*{6}{>{\centering\arraybackslash}m{1.1cm}}}
\specialrule{.12em}{.05em}{.05em}
Modality && \multicolumn{2}{c}{\footnotesize{Cataract Surgery}} & \multicolumn{2}{c}{\footnotesize{Laparoscopy Surgery}} & \multicolumn{1}{l}{\multirow{2}{*}{\footnotesize{Avg. IoU (\%)}}} \\ \cmidrule(lr){3-4}\cmidrule(lr){5-6}
Backbone &  Network & IoU (\%) & Dice (\%) & IoU (\%) & Dice (\%) \\ \specialrule{.12em}{.05em}{.05em}
\multirow{7}{*}{VGG16} & UNet+ & 45.79 & 56.73 & \textcolor{Blue}{57.74} & \textcolor{Blue}{70.29} & 51.76\\
 & scSENet & 45.74 & 56.19 & 56.14 & 69.08 & 50.94\\
 & FEDNet & 44.25 & 55.45 & 53.69 & 66.69 & 48.97\\
 & CE-Net & 41.72 & 53.04 & 48.91 & 62.34 & 45.31\\
 & CPFNet & \textcolor{Blue}{49.42} & \textcolor{Blue}{60.15} & 57.16 & 69.60 & \textcolor{Blue}{53.29}\\
 & UNetPP & 45.74 & 56.67 & 57.40 & 69.88 & 51.57\\
 & DeepPyramid+ & \textcolor{ForestGreen}{56.48} & \textcolor{ForestGreen}{66.40} & \textcolor{ForestGreen}{61.39} & \textcolor{ForestGreen}{73.09} & \textcolor{ForestGreen}{58.93}\\ \midrule
\multirow{8}{*}{ResNet34} 
 & scSENet & 50.00 & 60.91 & 50.71 & 63.88 & 50.35\\
 & FEDNet & 47.10 & 58.68 & 23.34 & 32.44 & 35.22\\
 & CE-Net & 16.46 & 26.52 & 30.73 & 44.48 & 23.59\\
 & CPFNet & 28.70 & 41.14 & 49.28 & 63.40 & 38.99\\
 & RAUNet & 43.36 & 55.36 & 1.62 & 2.97 & 22.49\\
 & BARNet & \textcolor{ForestGreen}{51.78} & \textcolor{ForestGreen}{63.09} & \textcolor{Blue}{51.09} & \textcolor{Blue}{64.70} & \textcolor{Blue}{51.43}\\
 & PAANet & \textcolor{Blue}{51.14} & \textcolor{Blue}{58.68} & 48.91 & 61.85 & 50.02\\
 & DeepPyramid+ & 49.11 & 59.72 & \textcolor{ForestGreen}{57.14} & \textcolor{ForestGreen}{69.80} & \textcolor{ForestGreen}{53.12}\\ \midrule
\multirow{3}{*}{ResNet50}
 & UPerNet & \textcolor{ForestGreen}{56.27} & \textcolor{ForestGreen}{66.93} & \textcolor{Blue}{56.08} & \textcolor{Blue}{68.70} & \textcolor{ForestGreen}{56.17}\\
 & DeepLabV3+ & 36.98 & 49.16 & 48.80 & 62.04 & 42.89\\
 & DeepPyramid+ & 49.28 & 59.90 & \textcolor{ForestGreen}{57.12} & \textcolor{ForestGreen}{70.06} & \textcolor{Blue}{53.20}\\

\specialrule{.12em}{.05em}{.05em}
\end{tabular}
      }
\end{table}

Table~\ref{tab:quantitative-disease} reports the segmentation performance of the proposed and state-of-the-art networks across three different modalities. DeepPyramid+ consistently demonstrates the highest average performance across all datasets with various backbones, while other methods, such as CPFNet,  exhibit varying performance with different backbones.
and 2.22\% compared to DeepPyramid+, respectively.
Besides, DeepPyramid+ achieves the best results with all three backbones for endometrial implants and prostate segmentation and the best results with ResNet34 and ResNet50 backbones for IRF segmentation in OCT.  
Considering instrument segmentation performance (Table~\ref{tab:quantitative-instruments}), DeepPyramid+ with VGG16 backbone shows more than 5.6\% gain in segmentation compared to CPFNet as its main alternative (58.93\% vs. 53.29\%). Across all backbones, DeepPyramid+ with VGG16 backbone shows more than 2.7\% higher performance compared to other methods. Besides, the best results for both datasets correspond to DeepPyramid+ with VGG16 backbone. Overall, DeepPyramid+ with our suggested backbone (VGG16) achieves the best segmentation performance in instrument and organ/disease segmentation.

\begin{table}[tb!]
\centering
\caption{Quantitative comparisons of cross-domain performance among DeepPyramid+ and state-of-the-art methods, with top two results shown in \textcolor{ForestGreen}{green} and \textcolor{Blue}{blue}, respectively.}
\label{tab:quantitative-domain-adaptation}
\resizebox{0.83\textwidth}{!}{%

\begin{tabular}{lm{1.3cm}*{2}{>{\centering\arraybackslash}m{1.3cm}}m{1.3cm}*{2}{>{\centering\arraybackslash}m{1.3cm}}}
\specialrule{.12em}{.05em}{.05em}
Modality & \multicolumn{3}{c}{\footnotesize{MRI}} & \multicolumn{3}{c}{\footnotesize{Cataract Surgery}}  \\ \cmidrule(lr){2-4}\cmidrule(lr){5-7}
Backbone &  Network & IoU (\%) & Dice (\%) & Network & IoU (\%) & Dice (\%) \\ \specialrule{.12em}{.05em}{.05em}
\multirow{3}{*}{VGG16} & CPFNet & \textcolor{Blue}{40.66} & \textcolor{Blue}{54.24} & UNet+ & \textcolor{Blue}{26.26} & \textcolor{Blue}{35.59}\\
& UNet++ & 36.30 & 49.30 & CPFNet & 25.14 & 34.04 \\
& DeepPyramid+ & \textcolor{ForestGreen}{44.43} & \textcolor{ForestGreen}{59.11} & DeepPyramid+ & \textcolor{ForestGreen}{42.93} & \textcolor{ForestGreen}{55.10}\\ \midrule
\multirow{3}{*}{ResNet34} & scSENet & \textcolor{Blue}{40.23} & \textcolor{Blue}{53.19} & BARNet & \textcolor{Blue}{20.22} & \textcolor{Blue}{29.31} \\
& FEDNet & 33.05 & 44.47 & PAANet & 14.01 & 20.48\\
& DeepPyramid+ & \textcolor{ForestGreen}{41.52} & \textcolor{ForestGreen}{56.14} & DeepPyramid+ & \textcolor{ForestGreen}{32.87} & \textcolor{ForestGreen}{43.96}\\ \midrule
\multirow{3}{*}{ResNet50} & UperNet & \textcolor{Blue}{38.45} & \textcolor{Blue}{51.78} & UPerNet & \textcolor{Blue}{28.40} & \textcolor{Blue}{38.10}\\
& DeepLabV3+ & 37.55 & 49.62 & DeepLabV3+ & 9.14 & 14.16 \\
& DeepPyramid+ & \textcolor{ForestGreen}{38.89} & \textcolor{ForestGreen}{53.00} & DeepPyramid+ & \textcolor{ForestGreen}{29.76} & \textcolor{ForestGreen}{40.54}\\

\specialrule{.12em}{.05em}{.05em}
\end{tabular}
      }
\end{table}

Table~\ref{tab:quantitative-domain-adaptation} compares the cross-domain segmentation performance of DeepPyramid+ and its best two alternatives for three backbones 
(considering single-domain results in Table~\ref{tab:quantitative-disease} and Table~\ref{tab:quantitative-instruments}). Overall, DeepPyramid+ consistently outperforms other methods across all backbones.
Considering the MRI dataset, DeepPyramid+ with VGG16 backbone shows more than 4.8\% gain in Dice compared to alternatives. 
For instrument segmentation in cataract surgery, DeepPyramid+ with the VGG16 backbone exhibits an impressive improvement of approximately 19.5\% in Dice score compared to CPFNet with the same backbone (55.10\% vs. 35.59\%), and a 17\% improvement compared to the best alternative across all backbones (55.10\% vs. 38.10\% achieved by UPerNet).
This exceptional performance in dealing with cross-domain distribution gaps \cite{ghamsarian2023domain} can be attributed to the effectiveness of the proposed modules in incorporating multi-scale local and global features. 

Table~\ref{tab:ablation} provides an ablation study of DeepPyramid+ components. The results suggest that both PVF and DPR modules contribute significantly to improvements in segmentation performance across all datasets. This impact is more prominent in the case of cataract surgery, where the addition of PVF and DPR modules lead to a 4.95\% and 4.72\% increase in the Dice coefficient, respectively.

\begin{table}[tb!]
\centering
\caption{Ablation study of DeepPyramid+ component across different datasets.}
\label{tab:ablation}
\resizebox{0.9\textwidth}{!}{%

\begin{tabular}{cc*{8}{>{\centering\arraybackslash}m{1.1cm}}}
\specialrule{.12em}{.05em}{.05em}
& & \multicolumn{2}{c}{\footnotesize{Endometriosis}} & \multicolumn{2}{c}{\footnotesize{MRI}} & \multicolumn{2}{c}{\footnotesize{Cataract Surgery}} & \multicolumn{2}{c}{\footnotesize{Laparoscopy Surgery}}\\ \cmidrule(lr){3-4}\cmidrule(lr){5-6}\cmidrule(lr){7-8}\cmidrule(lr){9-10}
PVF & DPR & IoU (\%) & Dice (\%) & IoU (\%) & Dice (\%) & IoU (\%) & Dice (\%) & IoU (\%) & Dice (\%)\\ \specialrule{.12em}{.05em}{.05em}

\XSolidBrush & \XSolidBrush &   51.02 & 64.94 & 72.44 & 82.30 & 45.79 & 56.73 & 57.74 & 70.29\\
\CheckmarkBold & \XSolidBrush  &  52.68 & 66.14 & 74.51 & 84.41 & 51.42 & 61.68 & 60.65 & 72.72\\
\CheckmarkBold & \CheckmarkBold &   53.22 & 66.37 & 76.02 & 85.36 & 56.48 & 66.40 & 61.39 & 73.09\\

\specialrule{.12em}{.05em}{.05em}
\end{tabular}

      }
\end{table}

\section{Conclusion}
\label{sec: Conclusion}
In recent years, considerable attention has been devoted to computerized medical image and surgical video analysis.
A reliable relevant-instance-segmentation approach is a prerequisite for a majority of these applications. 
In this paper, we introduce a novel network architecture for semantic segmentation that addresses the challenges encountered in medical image and surgical video segmentation. Our proposed architecture, DeepPyramid+, incorporates two innovative modules, namely ``Pyramid View Fusion" and ``Deformable Pyramid Reception".
Experimental results demonstrate the effectiveness of DeepPyramid+ in capturing object features in challenging scenarios, including shape and scale variation, reflection and blur degradation, blunt edges, and deformability, resulting in competitive performance in cross-domain segmentation compared to state-of-the-art networks. The ablation study validates the efficacy of the proposed modules in DeepPyramid+, showcasing their performance across diverse datasets.
The obtained promising results indicate the potential of DeepPyramid+ to enhance the precision in various computerized medical imaging and surgical video analysis applications.

% \section*{Acknowledgement}
% This work was funded by Haag-Streit Switzerland.

\section*{Declarations}

\noindent\textbf{\textit{Conflicts of interest. }}The authors declare that they have no conflicts of interest.

\vspace{0.5\baselineskip}
\noindent\textbf{\textit{Ethical approval. }}For this type of study, formal consent is not required.

\vspace{0.5\baselineskip}
\noindent\textbf{\textit{Informed consent. }}This article uses patient data from publicly available datasets.

\bibliography{bibtex}

\end{document}